# SIMULTANEOUS DEMPSTER-SHAFER CLUSTERING AND GRADUAL DETERMINATION OF NUMBER OF CLUSTERS USING A NEURAL NETWORK STRUCTURE

*Johan Schubert*

Department of Information System Technology
Division of Command and Control Warfare Technology
Defence Research Establishment
SE–172 90 Stockholm, Sweden
schubert@sto.foa.se

## ABSTRACT

In this paper we extend an earlier result within Dempster-Shafer theory ["Fast Dempster-Shafer Clustering Using a Neural Network Structure," in *Proc. Seventh Int. Conf. Information Processing and Management of Uncertainty in Knowledge-Based Systems* (IPMU'98)] where several pieces of evidence were clustered into a fixed number of clusters using a neural structure. This was done by minimizing a metaconflict function. We now develop a method for simultaneous clustering and determination of number of clusters during iteration in the neural structure. We let the output signals of neurons represent the degree to which a pieces of evidence belong to a corresponding cluster. From these we derive a probability distribution regarding the number of clusters, which gradually during the iteration is transformed into a determination of number of clusters. This gradual determination is fed back into the neural structure at each iteration to influence the clustering process.

## 1. INTRODUCTION

In this paper we develop a neural network structure for simultaneous clustering of evidence within Dempster-Shafer theory [1] and gradual determination of number of clusters. The clustering is done by minimizing a metaconflict function. The studied problem concerns the situation when we are reasoning with multiple events which should be handled independently. We use the clustering process to separate the evidence into clusters that will be handled separately.

In an earlier paper [2] we developed a method based on clustering with a neural network structure into a fixed number of clusters. We used the structure of the neural network, but no learning was done to set the weights of the network. Instead, all the weights were set directly by a method using the conflict in Dempster's rule as input. This clustering approach was a great improvement on computational complexity compared to a previous method based on iterative optimization [3–7], although its clustering performance was not equally good. In order to improve on clustering performance a hybrid of the two methods has also been developed [8].

Here, the idea of gradual determination of number of clusters is developed and integrated with the neural structure to run simultaneously with the clustering process. In [5] a methodology was developed for finding a posterior probability distribution concerned with the number of clusters. This was based on the final clustering result of an iterative optimization of the metaconflict function and partial specifications of nonspecific pieces of evidence, uncertain with respect to which events they were referring to [4]. Here, this approach is expanded to include a gradual determination of number of clusters. Instead of using the final clustering result and partial specifications, we use incremental clustering states during the iteration in the neural network, where we let an output signal of a neuron represent the degree to which a piece of evidence belongs to the corresponding cluster. This yields a probability distribution for the number of clusters. By using a change in entropy of the output signals in the neural structure during iteration, we can gradually transform the probability distribution into a determination of number of clusters. The result of this gradual determination is fed back into the neural network during the iteration to influence the clustering process. In the early stages of the iteration the number of clusters vary slowly with the change in the gradual determination of number of clusters and any changes in conflict in the clusters. In the latter stages, the iteration converges on a fixed number of clusters.

The idea to use a neural network for optimization was inspired by an approximate solution to the traveling salesman problem by Hopfield and Tank [9].

The clustering methodology developed over several papers was initially intended for preprocessing of intelligence information for situation analysis in antisubmarine warfare [10–12].

In section 2 we describe the problem at hand and in section 3 we continue with the neural structure for simultaneous clustering and gradual determination of number of clusters. First, we describe the approach to clustering with a neural structure. Secondly, we focus on the gradual determination of number of clusters and how this is integrated into the neural structure to run simultaneously with the clustering process. The computational and clustering performance is investigated in section 4. Finally, in section 5, we draw conclusions.

## 2. THE PROBLEM

If we receive several pieces of evidence about different and separate events and the pieces of evidence are mixed up, we want to arrange them according to which event they are referring to. Thus, we partition the set of all pieces of evidence $\chi$ into subsets where each subset refers to a particular event. In figure 1 these subsets are denoted by $\chi_i$ and the conflict when all pieces of evidence in $\chi_i$ are combined by Dempster's rule is denoted by $c_i$. Here, thirteen pieces of evidence are partitioned into four subsets. When the number of subsets is uncertain there will also be a "domain conflict" $c_0$ which is a conflict between the current hypothesis about the number of subsets and our prior belief. The partition is then simply an allocation of all pieces of evidence to the different events. Since these events do not have anything to do with each other, we will analyze them separately.

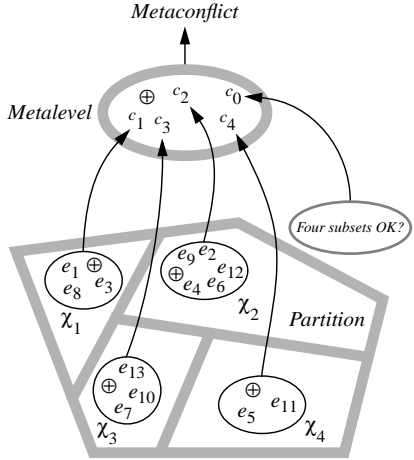

Fig 1. The conflict in each subset of the partition becomes a piece of evidence at the metalevel.

Now, if it is uncertain to which event some pieces of evidence is referring we have a problem. It could then be impossible to know directly if two different pieces of evidence are referring to the same event. We do not know if we should put them into the same subset or not. This problem is then a problem of organization. Evidence from different events that we want to analyze are unfortunately mixed up and we are facing a problem in separating them.

To solve this problem, we can use the conflict in Dempster's rule when all pieces of evidence within a subset are combined, as an indication of whether these pieces of evidence belong together. The higher this conflict is, the less credible that they do.

Let us create an additional piece of evidence for each subset with the proposition that this is not an "adequate partition". We have a simple frame of discernment on the metalevel $\Theta = \{AdP, \neg AdP\}$, where AdP is short for "adequate partition." Let the proposition take a value equal to the conflict of the combination within the subset,

$$m_{\chi_i}(\neg AdP) \triangleq \text{Conf}(\{e_j | e_j \in \chi_i\}).$$

These new pieces of evidence, one regarding each subset, reason about the partition of the original evidence. Just so we do not confuse them with the original evidence, let us call this evidence "metalevel evidence" and let us say that its combination and the analysis of that combination take place on the "metalevel," figure 1.

We establish [3] a criterion function of overall conflict called the metaconflict function for reasoning with multiple events. The metaconflict is derived as the plausibility that the partitioning is correct when the conflict in each subset is viewed as a piece of metalevel evidence against the partitioning of the set of evidence, $\chi$, into the subsets, $\chi_i$.

DEFINITION. *Let the* metaconflict function,

$$Mcf(r, e_1, e_2, \ldots, e_n) \triangleq 1 - (1 - c_0) \cdot \prod_{i=1}^{r}(1 - c_i),$$

*be the conflict against a partitioning of n pieces of evidence of the set* $\chi$ *into r disjoint subsets* $\chi_i$. *Here, $c_i$ is the conflict in subset i and $c_0$ is the conflict between r subsets and propositions about possible different number of subsets.*

We will use the minimizing of the metaconflict function as the method of partitioning the evidence into subsets representing the events. This method will also handle the situation when the number of events are uncertain.

### 3. NEURAL STRUCTURE

We will study a test problem where 31 pieces of evidence, all simple support functions with elements from $2^\Theta$, are clustered into an unknown number of clusters, where $\Theta = \{1, 2, 3, \ldots, 5\}$.

Since the elements are all the different subsets of the frame there is always a partition into five clusters with a global minimum to the metaconflict function equal to zero. If you put all evidence with the 1–element into $\chi_1$, of the remaining elements all those with the 2–element into $\chi_2$, etc., you get zero conflict in every cluster. The reason we choose a problem where the minimum metaconflict is zero is that it makes a good test example for evaluating performance.

When minimizing the metaconflict function using a neural structure we will choose an architecture that minimizes a sum. Thus, we have to make some change to the function that we want to minimize. If we take the logarithm of one minus the metaconflict function, we can change from minimizing Mcf to minimizing a sum.

Let us change the minimization as follows

$$\min Mcf = \min 1 - \prod_i (1 - c_i)$$
$$\max 1 - Mcf = \max \prod_i (1 - c_i)$$
$$\max \log(1 - Mcf) = \max \log \prod_i (1 - c_i)$$
$$= \max \sum_i \log(1 - c_i) = \min \sum_i -\log(1 - c_i)$$

where $-\log(1 - c_i) \in [0, \infty]$ is a weight.

Since the minimum of Mcf is obtained when the final sum is minimal, the minimization of the final sum yields the same result as a minimization of Mcf would have.

Thus, in the neural network we will not let the weights be directly dependent on the conflicts between different pieces of evidence but rather on $-\log(1 - c_{jk})$, where $c_{jk}$ is the conflict between the *j*th and *k*th piece of evidence;

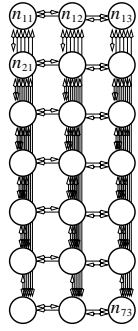

Fig 2. Neural network. Each column corresponds to a cluster and each row corresponds to a piece of evidence.

$$c_{jk} = \begin{cases} m_j \cdot m_k, & \text{conflict} \\ 0, & \text{no conflict.} \end{cases}$$

This is a slight simplification since the neural structure will now minimize a sum of $-\log(1 - c_{jk})$.

Let us study the calculations taking place in the neural network during an iteration. We use the same terminology as Hopfield and Tank [4] with input voltages as the weighted sum of input signals to a neuron, output voltages as the output signal from a neuron, and inhibition terms as negative weights.

For each neuron $n_{mn}$ we calculate an input voltage u as the weighted sum of all signals from row $m$ and column $n$ and from a domain term, figure 2.

This sum is the previous input voltage for the previous iteration of $n_{mn}$ plus a gain factor η times a sum of five terms. The first term is the sum of output voltages $V_{in}$ of all neurons of the same column as $n_{mn}$, weighted by a data-term inhibition dti times the weight of conflict $-\log(1 - c_{in})$ plus a global inhibition gi. The second term is the sum of output voltages $V_{mj}$ of all neurons of the same row as $n_{mn}$, weighted by a row-inhibition ri and the global inhibition. The third term is the gradual determination of number of clusters. For a neuron $n_{mn}$ in column $r$ we have the gradual determination factor $gd_t(|\chi| = r)$ weighted by a domain-inhibition Dti plus the global inhibition. The two last terms are an excitation bias eb minus the previous input voltage of $n_{mn}$.

Thus, the new input voltage to $n_{mn}$ at iteration $t + 1$ is

$$u_{mn}^{t+1} = u_{mn}^{t} + \eta \cdot \left( \sum_{i} [-\text{dti} \cdot \log(1 - c_{in}) + \text{gi}] \cdot V_{in} \right.$$
$$\left. + \sum_{j \neq n} [\text{ri} + \text{gi}] \cdot V_{mj} + [\text{Dti} + \text{gi}] \cdot gd_t(|\chi| = n) + \text{eb} - u_{mn}^{t} \right)$$

where $gd_t(|\chi| = n) \in [0, 1]$, see section 3.1.

We have used the following parameter settings: η = $10^{-5}$, dti = −2000, ri = −500, Dti = −2000, gi = −200, and the excitation bias eb = 1800.

From the new input voltage to $n_{mn}$ we can calculate a new output voltage of $n_{mn}$

$$V_{mn}^{t+1} = \frac{1}{2} \cdot \left( 1 + \tanh\left(\frac{u_{mn}^{t+1}}{u_0}\right) \right)$$

where tanh is the hyperbolic tangent, $u_0 = 0.02$, and $V_{mn}^{t+1} \in [0, 1]$.

Initially, before the iteration begins, each neuron is

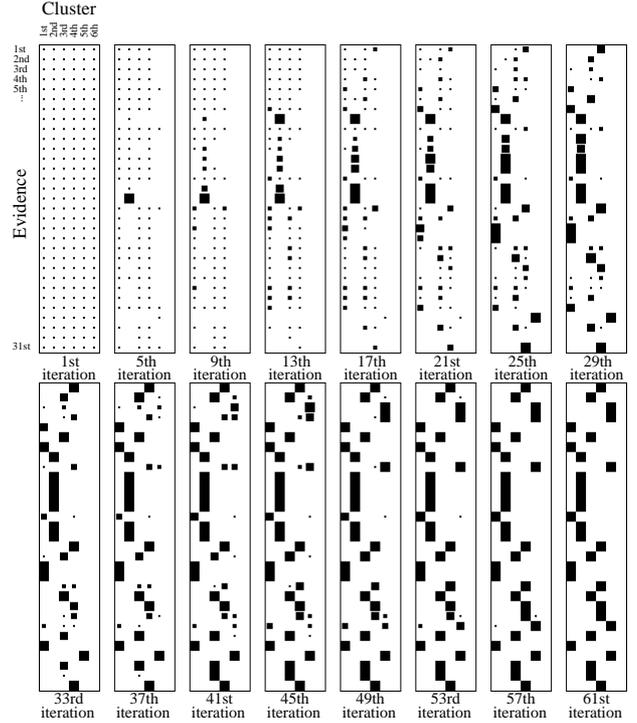

Fig 3. Sixteen different states (iterations) of a neural network with 186 neurons. From left to right: The convergence of clustering 31 pieces of evidence into an unknown number of clusters at every fourth iteration. In each snapshot of an iteration each of the six columns represent one possible cluster and each of the 31 rows represent one piece of evidence. The linear dimension of each square is proportional to the output voltage of the neuron and represent the degree to which a piece of evidence belongs to a cluster. In the final state each row has one output voltage of 1.0 and five output voltages of 0.0. A piece of evidence, represented by a row, is now clustered into the cluster where the output voltage is 1.0.

initiated with an input voltage of $u_{00}$ + noise where

$$u_{00} = u_0 \cdot \text{atanh}\left(\frac{2}{n} - 1\right)$$

and atanh is the hyperbolic arc tangent.

The initial input voltage is set at $u_{00} + \delta u$ where $\delta u$, the noise, is a random number chosen uniformly in the interval $-0.1 \cdot u_0 \leq \delta u \leq 0.1 \cdot u_0$.

In each iteration all new voltages are calculated from the results of the previous iteration. This continues until convergence is reached. As long as the weights of the neural network is symmetric convergence is always guaranteed.

In figure 3 the convergence of a neural network with 31 rows and six columns for clustering 31 pieces of evidence into a unknown number of clusters is shown.

After convergence is achieved the conflict within each cluster is calculated by combining those pieces of evidence for which the output voltage for the column is 1.0.

We now have a conflict for each subset and can calculate the overall metaconflict, Mcf, by the previous formula.

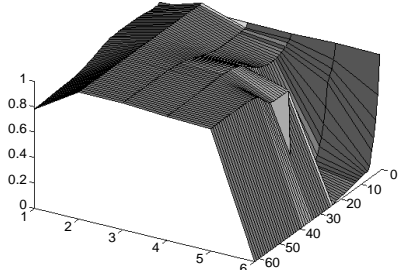

Fig 4. $m_{\chi_i}(\chi_i \in \chi)$ for all $\chi_i$ over 62 iterations.

## 3.1 Gradual determination of number of clusters

Making a gradual determination of the number of clusters is a process with several steps. First, we use the idea that each piece of evidence in a subset supports the existence of that subset to the degree that that piece of evidence supports anything at all [5]. During an iteration it is uncertain to which cluster a pieces of evidence belongs. Therefore, we will use every piece of evidence in each cluster but weighted by its output voltage for the cluster. All these weighted pieces of evidence in each cluster $\chi_i$ are then combined. The degree to which the result from this combination supports anything at all other than the entire frame is the degree to which these pieces of evidence taken together supports the existence of $\chi_i$. Thus, we have

$$m_{\chi_i}(\chi_i \in \chi) = 1 - \frac{1}{1-k} \cdot \prod_m (1 - V_{mn}^{t+1} + V_{mn}^{t+1} \cdot m_m(\Theta)),$$

$$m_{\chi_i}(\Theta) = \frac{1}{1-k} \cdot \prod_m (1 - V_{mn}^{t+1} + V_{mn}^{t+1} \cdot m_m(\Theta))$$

where $k$ is the conflict in Dempster's rule of the combination. Should nothing be supported by the evidence in some cluster that evidence is meaningless and can be thrown away and this particular cluster is not needed.

In figure 4 $m_{\chi_i}(\chi_i \in \chi)$ is calculated for all six possible clusters over 62 iterations. This is the same problem as in figure 3. We notice how the basic probability of the 6th clusters drops off very fast during the first few iterations and remains low for the remainder of the iterations. Also the basic probability of the 5th cluster drops off initially but comes back again during the iteration process to finish high.

These six pieces of evidence $m_{\chi_i}$, one regarding each subset, are then combined. We have

$$m_{\chi}((\wedge \chi^*) \in \chi) = \prod_{i|(\chi_i \in \chi^*)} m_{\chi_i}(\chi_i \in \chi) \cdot \prod_{j|\chi_j \notin \chi^*} m_{\chi_j}(\Theta),$$

$$m_{\chi}(\Theta) = \prod_{i=1}^{n} m_{\chi_i}(\Theta).$$

where $\chi^* \in 2^\chi$ and $\chi = \{\chi_1, \chi_2, ..., \chi_6\}$.

From this we can create a new type of evidence by exchanging all propositions in the previous ones that are conjunctions of $r$ terms, $\wedge \chi^*$, for one proposition in the new type of evidence that is on the form $|\chi| \geq r$. Here, $1 \leq r \leq 6$. The sum of probability of all conjunctions of length $r$ in the previous pieces of evidence is then awarded to the focal element in this new piece of evidence which supports the proposition that $|\chi| \geq r$. We get,

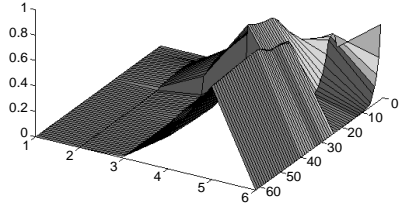

Fig 5. $m_\chi(|\chi| \geq r)$ for all $\chi_i$ over 62 iterations.

$$m_\chi(|\chi| \geq r) = \sum_{\chi^*||\chi^*|=r} m_\chi((\wedge \chi^*) \in \chi),$$

$$m_\chi(\Theta) = m_\chi(\Theta).$$

Taken as a whole this gives us an opinion about the probability of different numbers of subsets.

In figure 5 we see the basic probability $m_\chi(|\chi| \geq r)$ for different minimal number of clusters over the entire iteration. Note that we are no longer talking about individual clusters but of the derived support regarding the actual number of clusters.

These newly created pieces of evidence can now be combined with a prior probability distribution, $m(\cdot)$, from the problem specification. This is a domain dependent distribution. Without a prior probability distribution there is nothing to hold the clustering together and we could find as many clusters as pieces of evidence if that was allowed by the neural structure. In these trials we have chosen a distribution where $m(|\chi| = r) = p^{2 \cdot (r-1)}$, where $p$ is a constant. We get

$$m^*(|\chi| = r) = \frac{1}{1-k} \cdot m(|\chi| = r) \cdot \left( m_\chi(\Theta) + \sum_{j=1}^{i} m_\chi(|\chi| \geq j) \right)$$

where

$$k = \sum_{i=0}^{n-1} \sum_{j=i+1}^{n} m(|\chi| = r) \cdot m_\chi(|\chi| \geq j)$$

is the conflict in that final combination.

Thus, by viewing each piece of evidence in a subset as support for the existence of that subset we are able to derive a posterior probability distribution concerned with the question of how many subsets there are.

This distribution is plotted in figure 6. It is interesting to observe that the basic probabilities for propositions such as $m_\chi(|\chi| \geq r)$ has now been moved upwards when we derive the probabilities $m^*(|\chi| = r)$.

In figure 7 we see the same plot projected on the iteration—probability axes. We notice the initial drop off in the alternatives with five and especially with six clusters, as well as the early rise of the three and four cluster alternatives. In the first few iterations the six cluster alternative is still the preferable choice, but very quickly the preferable choice becomes five and then four clusters. After trying to cluster the evidence into four clusters the internal conflicts become too large and it is necessary to reclaim the fifth cluster. This is observed here in the dramatic shift around the 23 iteration. After that shift, the five cluster alternative remains preferable throughout the remainder of the iteration.

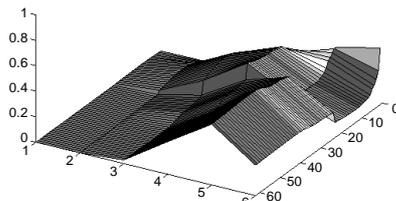

Fig 6. $m^*(|\chi| = r)$ for all $\chi_i$ over 62 iterations.

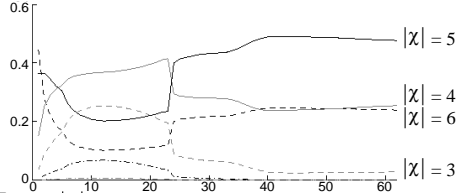

Fig 7. $m^*(|\chi| = r)$.

Sooner or later it becomes necessary to determine how many clusters there are. The basis for making such a determination is the probability distribution in figure 7. The obvious choice is that of the highest probability, and that is how we will decide in the final iteration, but we must avoid making an early determination. As seen in figure 7, an early choice based on maximum probability is likely to be false. However, neither can we wait until the final iteration to determine the number of clusters. We must give the neural iteration a good opportunity to converge on the right problem.

The idea is to accept the probability distribution in figure 7 but make a gradual determination of number of clusters throughout the iterative process that is not final until the last iteration. We do this by measuring how far we have traveled from the initial state until the final state when convergence is reached. Initially all output voltages of the neurons are scattered maximally among the different choices while in the final state there is no scattering at all. A normalized Shannon entropy is thus a good measure of how far we have traveled towards the final state, the entropy being zero at the final state. The entropy at iteration $t$ is calculated over all pieces of evidence and all clusters as

$$\text{Entropy}_t = -\sum_m \sum_n V^t_{mn} \log V^t_{mn}$$

where the normalized entropy $\alpha_t$ is $\text{Entropy}_t$ divided by $\text{Entropy}_0$. The normalized entropy is plotted in figure 8 as a solid line, it goes from 1 to 0 in 62 iterations.

The gradual determination is made by substituting the probability distribution in figures 6 and 7 with

$$gd_t(|\chi| = r) = 1 - \alpha_t + \alpha_t \cdot m^*(|\chi| = r),$$

**iff** $\forall s.\ m^*(|\chi| = r) \geq m^*(|\chi| = s),$

$gd_t(|\chi| = r) = \alpha_t \cdot m^*(|\chi| = s)$, otherwise.

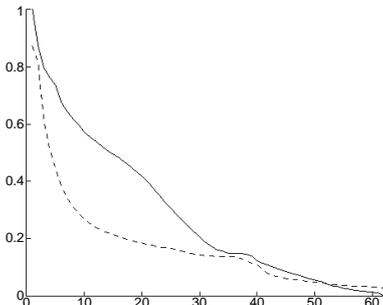

Fig 8. Normalized entropy $\alpha_t$, solid line, and metaconflict, dashed line, over 62 iterations.

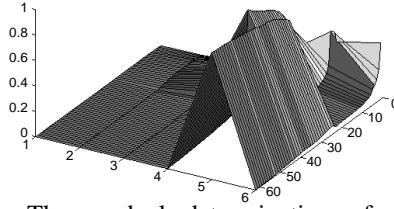

Fig 9. The gradual determination of number of clusters.

In figure 9 the gradual determination of the posterior probability distribution in figures 6 is plotted. In its first iteration they are identical, gradually the determination takes place and in the final iteration an exact determination of five clusters is made.

From figure 9 we observe that even though the exact determination of five clusters does not take place until the final iteration, the situation is pretty clear long before. At the 32nd and 44th iterations $gd_{32}(|\chi| = 5) = 0.901$ and $gd_{44}(|\chi| = 5) = 0.953$, respectively, giving the clustering process ample time to converge on the right number of clusters. Also, in figure 3 we noticed a good clustering taking place in the fifth cluster from the 37th iteration, giving the process some 25 additional iterations to converge.

## 4. PERFORMANCE

In this section we will compare the clustering performance and computational complexity of the simultaneous clustering and gradual determination of numbers of clusters presented in this paper with clustering into a known number of five clusters which was done in [2]. The comparison is done using the previous described 31 pieces of evidence with different random basic probability assignments over ten runs.

In figure 8 the metaconflict over 62 iterations of one of the runs resulting in five clusters is shown, dashed line. In figure 10 the conflict per cluster is shown. We notice the tendency that the conflict drops off in one cluster at a time. Quickly the conflict drops off in cluster six and then in five. Initially, there is a very high conflict in cluster two but it also, drops off within the first ten iterations. The remainder of the iteration concerns mainly clusters one, three and four. Cluster three holds out until the middle of the iteration but towards the end most of the remaining conflict is situated in cluster one.

In table 1 we find a somewhat higher computation time for the new approach, as would be expected, since we have two different converging processes running simultaneously.

Table 1: Computation time and iterations (mean).

| # Clusters | 5 | unknown |
|---|---|---|
| Neural structure | | |
| time (s) | 27.3 | 37.9 |
| iterations | 65.2 | 78.9 |

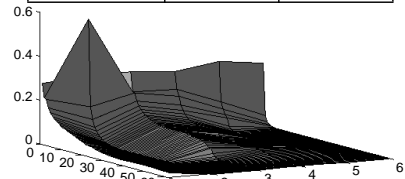

Fig 10. Conflict per cluster during the iteration.

If we consider that the clustering into a known number of clusters has to be performed for many possible different numbers of clusters one after the other, this can be a significant saving of time.

In the ten runs of clustering 31 pieces of evidence into an unknown number of clusters four of these runs ended up using five clusters, five runs used six clusters and one run used only four clusters. The use of six clusters in some of the runs is the result of a below average clustering performance. If the number of clusters had been fixed to five, we would have seen a higher than average conflict in these runs. Here, this is "interpreted" as a need for an additional cluster. Since the propositions of the evidence are the 31 different subsets of the frame $\Theta = \{1, 2, 3, ..., 5\}$, it will never be possible to find a conflict free partitioning of the evidence into four clusters. The reason why such a partition occurred once, is that the basic probability assignments attached to the propositions are random number between 0 and 1. If two propositions are in conflict and their assignments are close to zero the numeric conflict is still a small number and might be found acceptable compared to the use of an additional cluster. If all assignments had been 1.0 we could never have had four clusters in this test.

The clustering performance is somewhat difficult to measure because of the different number of subsets. A poor clustering might yield more clusters than a good one with a lower conflict per cluster. For this reason we will compare the performance of those runs that resulted in five clusters when clustering into a unknown number of clusters with clustering into a known five clusters. We have four runs resulting in five clusters. They are considered the four best runs out of ten. We compare them to the four best runs out of ten when clustering into a known five clusters. We also compare the four runs with the same set of random assignments when the number of clusters is known to be five, table 2.

Table 2: Metaconflict.

| # Clusters | 5 | unknown |
|---|---|---|
| 4 best runs of 10 | | |
| best of 4 | 0 | 0.024 |
| mean of 4 | 0.018 | 0.060 |
| 4 runs with same assignments | | |
| best of 4 | 0.011 | – |
| mean of 4 | 0.049 | – |

As expected, the conflict of the new approach is higher, but the actual numbers are quite small. The best criteria of a good clustering performance is the conflict per cluster and piece of evidence. These are tabulated in table 3. For example, we find a mean conflict per cluster and piece of evidence of only 2‰ in the case of clustering into an unknown number of clusters. This can be compared with the average numeric conflict of 25% between two conflicting pieces of evidence.

Table 3: Mean metaconflict per cluster and evidence.

| # Clusters | 5 | unknown |
|---|---|---|
| 4 best of 10 runs | | |
| / cluster | 0.0036 | 0.0122 |
| / evidence | 0.0006 | 0.0020 |
| 4 runs with same assignments | | |
| / cluster | 0.0100 | – |
| / evidence | 0.0016 | – |

## 5. CONCLUSIONS

We have demonstrated that it is possible to use a neural network structure to perform simultaneous clustering of Dempster-Shafer evidence with a gradual determination of the number of clusters when this is unknown. We found the computational and clustering performance to be almost as good as when clustering into a fixed number of clusters. This approach is advantageous, since now only one clustering process has to be performed.

## REFERENCES

[1] G. Shafer, A Mathematical Theory of Evidence, Princeton University Press, Princeton, 1976.
[2] J. Schubert, "Fast Dempster-Shafer Clustering Using a Neural Network Structure," in Proc. Seventh Int. Conf. Information Processing and Management of Uncertainty in Knowledge-Based Systems (IPMU'98), pp. 1438–1445, Université de La Sorbonne, Paris, France, 6–10 July 1998, Editions EDK, Paris, 1998.
[3] J. Schubert, "On Nonspecific Evidence," Int. J. Intell. Syst., Vol 8, pp. 711–725.
[4] J. Schubert, "Specifying Nonspecific Evidence," Int. J. Intell. Syst., Vol 11, pp. 525–563.
[5] J. Schubert, "Finding a Posterior Domain Probability Distribution by Specifying Nonspecific Evidence," Int. J. Uncertainty, Fuzziness and Knowledge-Based Syst., Vol 3, pp. 163–185.
[6] J. Schubert, "Cluster-based Specification Techniques in Dempster-Shafer Theory," in Symbolic and Quantitative Approaches to Reasoning and Uncertainty, Proc. European Conf. (ECSQARU'95), pp. 395–404, University of Fribourg, Switzerland, 3–5 July 1995, Springer-Verlag (LNAI 946), Berlin, 1995.
[7] J. Schubert, "Creating Prototypes for Fast Classification in Dempster-Shafer Clustering," in Qualitative and Quantitative Practical Reasoning, Proc. First Int. Joint Conf. (ECSQARU-FAPR'97), pp. 525–535, Bad Honnef, Germany, 9–12 June 1997, Springer-Verlag (LNAI 1244), Berlin, 1997.
[8] J. Schubert, "A Neural Network and Iterative Optimization Hybrid for Dempster-Shafer Clustering," in Proc. EuroFusion98 Int. Conf. Data Fusion, pp. 29–36, Great Malvern, UK, 6–7 October 1998.
[9] J.J. Hopfield, and D.W. Tank, "'Neural' Computation of Decisions in Optimization Problems," Biol. Cybern., Vol 52, pp. 141–152.
[10] U. Bergsten, J. Schubert, and P. Svensson, "Applying Data Mining and Machine Learning Techniques to Submarine Intelligence Analysis," in Proc. Third Int. Conf. on Knowledge Discovery and Data Mining (KDD'97), pp. 127–130, Newport Beach, USA, 14–17 August 1997, The AAAI Press, Menlo Park, 1997.
[11] J. Schubert, Cluster-based Specification Techniques in Dempster-Shafer Theory for an Evidential Intelligence Analysis of Multiple Target Tracks, Ph.D. Thesis, TRITA–NA–9410, Royal Institute of Technology, Sweden, 1994, ISRN KTH/NA/R--94/10--SE, ISSN 0348-2952, ISBN 91-7170-801-4.
[12] J. Schubert, "Cluster-based Specification Techniques in Dempster-Shafer Theory for an Evidential Intelligence Analysis of Multiple Target Tracks (Thesis Abstract)," AI Communications, Vol 8, pp. 1070–110.